# SalamNET at SemEval-2020 Task 12: Deep Learning Approach for Arabic Offensive Language Detection

**Fatemah Husain**
Kuwait University / State of Kuwait
f.husain@ku.edu.kw

**Jooyeon Lee**[*]
George Mason University / USA
jlee252@masonlive.gmu.edu

**Sam Henry**
George Mason University / USA
shenry20@masonlive.gmu.edu

**Özlem Uzuner**
George Mason University / USA
ouzuner@masonlive.gmu.edu

## Abstract

This paper describes SalamNET, an Arabic offensive language detection system that has been submitted to SemEval 2020 shared task 12: Multilingual Offensive Language Identification in Social Media. Our approach focuses on applying multiple deep learning models and conducting in depth error analysis of results to provide system implications for future development considerations. To pursue our goal, a Recurrent Neural Network (RNN), a Gated Recurrent Unit (GRU), and Long-Short Term Memory (LSTM) models with different design architectures have been developed and evaluated. The SalamNET, a Bi-directional Gated Recurrent Unit (Bi-GRU) based model, reports a macro-F1 score of 0.83.

## 1 Introduction

Online offensive language is a critical problem that is threatening the safety and well-being of society. Previous studies demonstrate the connection between different forms of online offensive language, such as hate speech, cyberbullying, and real-world violence (Sap et al., 2019). The attack on the Tree of Life synagogue on October 2018 in Pittsburgh, which was described as the deadliest attack in the Jewish community in the US, shows the severe effects of online hate speech as the shooter had an online profile full of hate speech language (Keyser, 2018). Given the significance of the impact online offensive language can have, further investigation of the problem is warranted. We hypothesize that developing a Natural Language Processing (NLP) approach with language specific features can help in detecting offensive language to stop these attacks. While there are several studies on English offensive language detection, very few studies are available for the Arabic language. The Arabic language has specific language characteristics and features that require distinct system development (Mubarak et al., 2020). The main goal of this paper is to develop a system for online Arabic offensive language detection.

We participated in SemEval 2020 shared task 12: Multilingual Offensive Language Identification in Social Media with two submissions, a Logistic Regression (LR) based model and a Recurrent Neural Network (RNN) based model. We evaluate the performance of both models through a detailed error analysis to enhance the performance until we arrive to the SalamNET system for online Arabic offensive language detection from Twitter. We explore multiple deep learning models, including an RNN, a Gated Recurrent Unit (GRU), a Bi-directional Gated

---

[*]Second first author

Recurrent Unit (Bi-GRU), Long-Short Term Memory (LSTM), and a Bi-directional Long-Short Term Memory (Bi-LSTM) models separately. Based on our findings, the highest macro-F1 score among all models is 0.83, which was reported by the SalamNET that consists of a Bi-GRU based model with Term Frequency-Inverse Document Frequency (TF-IDF) features.

## 2 Related Work

During recent years, the topic of offensive language detection has become an attractive topic for Arabic NLP researchers. Various forms of Arabic offensive language have been studied. This includes hate speech (Albadi et al., 2018; Albadi et al., 2019; Chowdhury et al., 2019), cyberbullying (Haidar et al., 2017; Haidar et al., 2018; Haidar et al., 2019), adult content (Alshehri et al., 2018; Abozinadah and Jones, 2017; Mubarak et al., 2017), and the broader category of offensive and abusive online language (Alakrot et al., 2018b; Mohaouchane et al., 2019; Mubarak et al., 2020; Husain, 2020).

Albadi, Kurdi, and Mishra (2018; 2019) study religious hate speech in Twitter data for the Arabic language. They use character n-gram features and AraVec word embeddings with multiple classifiers including LR, Support Vector Machine (SVM), and GRU. The GRU-based models acheived their highest F1 score of 0.77 using AraVec. Chowdhury et al. (2019) use the same dataset to investigate the effects of community interaction and social representations in detecting religious hate speech. They use a deep learning approach and study multiple features, such as word embedding, node embedding, sentence representation, and character n-gram features. They develop several deep learning models including GRU, LSTM, Bi-LSTM, Convolutional Neural Network (CNN), and Bi-GRU. They also combine multiple models using self-attention and Node2Vec criteria. Results show the highest accuracy score of 0.81, obtained by a combined model that consists of Bi-GRU and CNN using Node2Vec features. Meanwhile, the best F1 score, recall, and precision were achieved by the combined model that included LSTM and CNN using Node2Vec feature; 0.89, 0.78, and 0.86, respectively.

Mohaouchane, Mourhir, and Nikolov (2019) explored multiple deep learning models to classify Arabic offensive language with Arabic Youtube comments collected by (Alakrot et al., 2018a). They used AraVec word embedding with CNN, Bi-LSTM, Bi-LSTM with attention mechanism, and combined CNN and LSTM models. Their results demonstrated an overall better performance using a CNN with which they achieved an accuracy of 0.87, precision of 0.86, and F1 of 0.84. However, their combined CNN-LSTM model showed the best recall score of 0.83 versus 0.82 with the CNN model.

A more recent study by Mubarak et al. (2020) investigated Arabic offensive language related to vulgar language and hate speech for Twitter data using multiple classifiers and word embeddings. Their system includes LR, SVM, Decision Tree, Random Forest, Gaussian Naive Bayes, Perceptron, AdaBoost, and Gradient Boosting classifiers. In addition to applying FastText (Bojanowski et al., 2016), AraVec, Mazajak (Abu Farha and Magdy, 2019), and BERT$_{\text{base multilingual}}$ (Devlin et al., 2019) for word embeddings. Results showed the best F1 score of 0.79 for the SVM-based model with Mazajak word embeddings.

## 3 Methodology

In general, the SalamNET has a similar system pipeline to traditional classification systems. Steps are sequential, starting with text preprocessing, followed by feature engineering, classification, and lastly system evaluation. During the classification model phase, the system employs multiple classifiers to investigate the differences in performance among the different models.

### 3.1 Dataset

We use the Arabic OffensEval 2020 dataset (Mubarak et al., 2020; Zampieri et al., 2020) for training and evaluation. The dataset consists of 10,000 tweets labelled as either offensive or not offensive. The distribution of the classes is highly imbalanced, and out of the total tweets only

1,900 are labeled as offensive. We follow the same data split as the SemEval competition for which the data is split, 7,000 tweets for training, 1,000 tweets for development, and 2,000 tweets for testing datasets.

## 3.2 Preprocessing

We adopt the same preprocessing approach from Husain (2020). For this, Husain (2020) defines seven intensive preprocessing steps; 1) emoji and emoticon conversion to textual label that describe the content of them; 2) letter normalization that converts multiple letters forms to one form such as Alif (أ،آ،إ) to (ا), Alif Maqsura (ئ،ي) to (ى), and Ta Marbouta (ة) to (ه), and reduces repeated letters more than two times within a word to two times only; 3) dialect normalization to convert variation in nouns among dialects to their Modern Standard Arabic (MSA) forms; 4) hyponym conversion to hypernym such as mapping multiple animal names to the word 'animal'; 5) hashtags segmentation to remove the '#' symbol and replace '_' by a space; 6) miscellaneous cleaning process such as removing numbers, HTML tags, more than one space, special symbols, stopwords, and diacritics; 7) upsampling minority class, however, this step is not adopted to the final pipeline because it demonstrates negative effect on the performance of the classifier.

## 3.3 Feature Engineering

In our experiments we use two types of features, TF-IDF and the AraVec word embedding model. The first model we used is a simple TF-IDF model to calculate the vector of each word. TF-IDF vectors are commonly used in text classification. AraVec is an open source pre-trained word embedding for the Arabic language (Soliman et al., 2017) which has been used for detecting offensive language in Arabic YouTube comments (Mohaouchane et al., 2019) and for detecting religious hate speech on Twitter (Albadi et al., 2018). AraVec provides multiple models for generating word embeddings based on two main factors: (1) the technique used in building the embeddings, which is either skip-gram or Continuous Bag-Of-Word (CBOW), and (2) the corpus on which the embeddings are trained, either a Twitter corpus, a World Wide Web pages corpus, or an Arabic Wikipedia articles corpus that are collected by et al. (2017). We used AraVec word embedding trained on Twitter corpus and CBOW model.

## 3.4 Classification Models

### 3.4.1 Baseline Models

Our baseline model is an LR-based model. We experimented with using two features as input. (1) A character-based TF-IDF feature with 2 to 5 characters, and (2) an AraVec word embedding. The LR model is implemented using the Scikit-learn library of python.

### 3.4.2 Deep Learning Models

We compare the baseline model to several different deep learning architectures. These architectures include an RNN, a GRU, a Bi-GRU, an LSTM, and a Bi-LSTM. We optimized hyper-parameters for each architecture using grid searches. This optimization includes varying the dropout percentage of each architecture from 0.25 to 0.99, varying the number of hidden layers from 1 to 2, and varying the number of neurons for each layer from 50 to 300. We report results for each model using the hyper-parameters that yielded the highest accuracy. After the extensive search, we set epochs to 50 and drop outs to 0.5 for all models, hidden unit size to 300 and hidden layers of 2 for RNN, hidden vector size of 100, layers of 1 for GRU, BI-GRU, BI-LSTM and LSTM. The experimental hardware platform was the Intel Xeon E3 (32G memory, GTX 1080 Ti). The experimental software platform was the Ubuntu 17.10 operating system. The Keras library and the Scikit-learn library of Python were used to build the models and for the comparative experiments.

| Model | Feature | Precision | Recall | Macro-F1 | Weighted-F1 |
|---|---|---|---|---|---|
| LR | TF-IDF | **0.89** | 0.67 | 0.71 | 0.84 |
| LR | AraVec | 0.84 | **0.78** | **0.81** | **0.88** |

Table 1: Performance Evaluation of Baseline Models

| Model | Feature | Precision | Recall | Macro-F1 | Weighted-F1 |
|---|---|---|---|---|---|
| LSTM | TF-IDF | 0.73 | 0.74 | 0.73 | 0.73 |
| LSTM | AraVec | 0.68 | 0.72 | 0.70 | 0.70 |
| Bi-LSTM | TF-IDF | 0.78 | 0.78 | 0.78 | 0.78 |
| Bi-LSTM | AraVec | 0.69 | 0.72 | 0.70 | 0.71 |
| RNN | TF-IDF | 0.79 | **0.80** | 0.80 | 0.79 |
| RNN | AraVec | 0.68 | 0.71 | 0.69 | 0.68 |
| GRU | TF-IDF | 0.75 | 0.83 | 0.78 | 0.77 |
| GRU | AraVec | 0.70 | 0.71 | 0.70 | 0.70 |
| Bi-GRU | TF-IDF | **0.87** | 0.79 | **0.83** | **0.84** |
| Bi-GRU | AraVec | 0.68 | 0.71 | 0.69 | 0.68 |

Table 2: Performance Evaluation of Neural Network Models

## 4 Results

We use multiple performance evaluation metrics to accurately evaluate the behavior of our models, including precision, recall, macro F-measure, and weighted F1-Measure. We apply 10-fold cross validation in evaluating the performance of the deep learning models. Table 1 shows the results of the baseline models, and table 2 shows for the results of the deep learning models.

Macro-F1 is the primary evaluation metric of the shared task. The total number of participants was 53 and the highest achieved macro-F1 score 0.901. Our LR based baseline model with AraVec achieved a macro-F1 of 0.81 and ranked 34th, and our Bi-GRU based model with TF-IDF achieved a macro-F1 of 0.80 and ranked 38th. We explored more deep learning models to improve our results. The highest achieved macro-F1 among all classifiers; RNN, LSTM, Bi-LSTM, GRU, and Bi-GRU; is 0.83, which was generated by Bi-GRU based model with TF-IDF feature, while other models showed advantages in terms of macro-F1. We name the best performing system SalamNET, which consists of a Bi-GRU architecture with TF-IDF features.

These results indicate that for the LR-based model, AraVec word embeddings give better results than TF-IDF features. However, for all the deep learning models TF-IDF features give better performance than AraVec word embeddings.

## 5 Discussion and Error Analysis

To perform an error analysis we examined the predictions of the 2,000 tweets in the test dataset. From this, we retrieved the set of tweets that were misclassified by all deep learning models and explored their content. In total, 390 offensive tweets were labeled as not offensive, and 11 non-offensive tweets were labeled as offensive by all deep learning models. This finding is related to the class imbalanced distribution in the dataset, only 19% of the entire dataset represents offensive samples. The following are some sample of offensive tweets that were predicted as non-offensive by all deep learning models. In these examples, we show content before preprocessing for better representation:

1. وجمهور الاهلى لن ينسى موقفكم يا اتحاد العار يا مرتزقة يا مرتشين والحساب قريبا وياريت ماتتمسحش فى جمهور الاهلى عشان الكلاب اللى زيك جمهور الاهلى داس عليهم بالجزمة خلاص لان ده مقامك يا قذر

2. لا لحكم العسكر...لا لحكم العسكر لا للسيسى...لا لابن عوف... لا ولا وألف لا لالحكم العسكررررارحل يا بلحه يا ابن ملكيه ا...

3. الحرية لحازم عبد العظيم الحرية يا ولاد الجزم يا ظلمة. الحرية لحازم عبد العظيم

Most of the misclassified samples were offensive tweets in the Egyptian dialect, included the names of famous figures, and often related to sports (e.g. included team names, famous players, fans). Even though our preprocessing phase is identical to Husain (2020), which reports significant impacts on the performance of hate speech detection in the Arabic language by converting hyponym to hypernym, specifically for animal names, during the preprocessing. We noticed some misclassified tweets with animal names. Thus, further analysis related to how animal names are preprocessed should be considered in future works.

Further analysis showed that some tweets were misclassified by all AraVec based models, but correctly classified by most TF-IDF based models. For example, "يا مجرم يا خبيث أنت وأمثالك يا قذر". Another example "عميل لليهود كافة، وللنظام الاجرام بقطر الذي يستخدمك وأمثالك لمخططاتهم. الاحرام يستشري في دمائكم. تفضل يا بجا يا طاقيه هالسؤال تروح توجهه لأمن المطار يا بقردي يا جاهل اذا ممنوع من دخول المملكه كيف انختم على جوازه" is "هلالي غبي". Most of these tweets are relatively long in term of the number of words, and rarely include repetitions in words or letters. This indicates that TF-IDF models may do a better job of modeling tweets that use a large vocabulary.

Similarly, we analyzed tweets that were classified correctly by all AraVec based models but misclassified by most of TF-IDF based models. Some examples are:"يارب ياقوي يا عزيز يا حي يا قيوم يا عساني نبقى يا عمري حبايب وحبنا" and "يا جبار يامتكبر يا مهين أسألك بعزك الذي لا يرام وملكك الذي لا يضام يكبر معانا". The first tweet consists of a prayer and it is written in the Modern Standard Arabic (MSA) form of the Arabic language. The second example is written in Gulf dialect. Both are not offensive. The dataset contains multiple forms of the Arabic language, and we expected that some forms of the Arabic language would be underrepresented. Most of Twitter content is written in dialectal Arabic meaning that MSA is underrepresented, thereby negatively affecting classifier performance. Although we considered this situation in our design by converting some dialectal nouns to MSA during the preprocessing phase, this analysis shows that we may need to further improve our mechanism.

In general, we found that each classifier behaves differently and complementary. The best performing classifiers are sometimes not able to detect the true class of a sample, for which other classifiers were able to. For instance, this tweet "ده وقت الحب والانتماءوهما بكاني زملكاوي اه يا تشرت العمر يا ايض زملكاوي لآخر يوم في عمري كل الدعم لرجالة الزمالك معاكيازمالك" is classified correctly by all classifiers with all features except the Bi-GRU based model with TF-IDF feature, which reports the highest macro-F1 score. Thus, providing a mechanism to combine results from multiple models can improve the model.

## 5.1 Further Improvements

Based on the results of this study we recommend the following potential improvements for future systems focused on online offensive language detection for the Arabic language:

- Named Entity Recognition (NER) features may be helpful. This is because we noticed a common pattern of incorrect classifications for tweets containing mentions of famous figures.

- Balance the number of samples from offensive and not offensive classes to improve the behavior of the classifiers and reduce misclassifications related to under representation of minority class.

- Analyze the categories of animals based on the classes, offensive and not offensive. This can support extracting semantic relationships to differentiate between the offensive and not offensive animal categories rather than treating all animals on the same mechanism.

- Investigate the relationship between religious content and offensive content. While many non-offensive tweets consisted of prayers or verses from the Qura'an, some offensive tweets referred to specific beliefs, specific believers, or famous religious figures within an offensive context. Thus, better knowledge of this relationship could enrich the feature extraction phase.

- Generate a greater coverage of less common Arabic dialects. This improves the scalability of the model and ensures it covers a broader dialect with accurate prediction. Offensive language is a culture-specific problem and differs widely from one culture or sub-culture to another; samples from one dialect might not be representative to another dialect.

- Merge the results from multiple classifiers. We observed that predictions among classifiers were often complementary, and merging their predictions could improve performance.

# 6  Conclusion

This paper presents the SalamNet system for Arabic offensive language detection. We explored multiple deep learning models; such as RNN, LSTM, GRU, Bi-LSTM, and Bi-GRU; using TF-IDF feature and AraVec word embeddings. Our results show the best macro-F1 score of 0.83 by the Bi-GRU based classifier using the TF-IDF feature, which represents the SalamNET system. AraVec word embeddings increased performance over TF-IDF features when using traditional machine learning classifiers (such as LR), but not with deep learning classifiers. In our analysis, we identified future directions, including investigating features related to the NER, using animal categories features, investigating religion and belief related content, and expanding sample coverage to more dialects.

## References

Ehab A. Abozinadah and James H. Jones. 2017. A statistical learning approach to detect abusive twitter accounts. In *Proceedings of the International Conference on Compute and Data Analysis*, ICCDA '17, page 6–13, New York, NY, USA. Association for Computing Machinery.

Ibrahim Abu Farha and Walid Magdy. 2019. Mazajak: An online Arabic sentiment analyser. In *Proceedings of the Fourth Arabic Natural Language Processing Workshop*, pages 192–198, Florence, Italy, August. Association for Computational Linguistics.

Azalden Alakrot, Liam Murray, and Nikola Nikolov. 2018a. Dataset construction for the detection of anti-social behaviour in online communication in arabic. *Procedia Computer Science*, 142:174–181, 11.

Azalden Alakrot, Liam Murray, and Nikola S. Nikolov. 2018b. Dataset construction for the detection of anti-social behaviour in online communication in arabic. *Procedia Computer Science*, 142:174–181.

Nuha Albadi, Maram Kurdi, and Shivakant Mishra. 2018. Are they our brothers? analysis and detection of religious hate speech in the arabic twittersphere. In *2018 IEEE/ACM International Conference on Advances in Social Networks Analysis and Mining (ASONAM)*, pages 69–76, August.

Nuha Albadi, Maram Kurdi, and Shivakant Mishra. 2019. Investigating the effect of combining GRU neural networks with handcrafted features for religious hatred detection on arabic twitter space. *Social Network Analysis and Mining*, 9(41):1–19, August.

Ali Alshehri, El Moatez Billah Nagoudi, Hassan Alhuzali, and Muhammad Abdul-Mageed. 2018. Think before your click: Data and models for adult content in arabic twitter. In *Second Workshop on Text Analytics for Cybersecurity and Online Safety (TA-COS 2018)*. European Language Resources Association (ELRA), May.

Piotr Bojanowski, Edouard Grave, Armand Joulin, and Tomas Mikolov. 2016. Enriching word vectors with subword information. cite arxiv:1607.04606Comment: Accepted to TACL. The two first authors contributed equally.

Arijit Ghosh Chowdhury, Aniket Didolkar, Ramit Sawhney, and Rajiv Ratn Shah. 2019. ARHNet - leveraging community interaction for detection of religious hate speech in Arabic. In *Proceedings of the 57th Annual Meeting of the Association for Computational Linguistics: Student Research Workshop*, pages 273–280, Florence, Italy, July. Association for Computational Linguistics.

Jacob Devlin, Ming-Wei Chang, Kenton Lee, and Kristina Toutanova. 2019. BERT: Pre-training of deep bidirectional transformers for language understanding. In *Proceedings of the 2019 Conference of the North American Chapter of the Association for Computational Linguistics: Human Language Technologies, Volume 1 (Long and Short Papers)*, pages 4171–4186, Minneapolis, Minnesota, June. Association for Computational Linguistics.

Batoul Haidar, Maroun Chamoun, and Ahmed Serhrouchni. 2017. Multilingual cyberbullying detection system: Detecting cyberbullying in arabic content. In *2017 1st Cyber Security in Networking Conference (CSNet)*, pages 1–8, October.

Batoul Haidar, Maroun Chamoun, and Ahmed Serhrouchni. 2018. Arabic cyberbullying detection: Using deep learning. In *2018 7th International Conference on Computer and Communication Engineering (ICCCE)*, pages 284–289, September.

Batoul Haidar, Maroun Chamoun, and Ahmed Serhrouchni. 2019. Arabic cyberbullying detection: Enhancing performance by using ensemble machine learning. In *2019 International Conference on Internet of Things (iThings) and IEEE Green Computing and Communications (GreenCom) and IEEE Cyber, Physical and Social Computing (CPSCom) and IEEE Smart Data (SmartData)*, pages 323–327, July.

Fatemah Husain. 2020. OSACT4 shared task on offensive language detection: Intensive preprocessing based approach. In *The 4th Workshop on Open-Source Arabic Corpora and Processing Tools (OSACT4). Proceedings of the Twelfth International Conference on Language Resources and Evaluation (LREC 2020)*, Marseille, France, May.

Zachary Keyser. 2018. Gab, the 'alt-right twitter' used by antisemitic pittsburgh shooter. *The Jerusalem Post*, October.

Hanane Mohaouchane, Asmaa Mourhir, and Nikola Nikolov. 2019. Detecting offensive language on arabic social media using deep learning. In *2019 Sixth International Conference on Social Networks Analysis, Management and Security (SNAMS)*, pages 466–471, October.

Hamdy Mubarak, Kareem Darwish, and Walid Magdy. 2017. Abusive language detection on Arabic social media. In *Proceedings of the First Workshop on Abusive Language Online*, pages 52–56, Vancouver, BC, Canada, August. Association for Computational Linguistics.

Hamdy Mubarak, Ammar Rashed, Kareem Darwish, Younes Samih, and Ahmed Abdelali. 2020. Arabic offensive language on twitter: Analysis and experiments. *arXiv preprint arXiv:2004.02192*.

Maarten Sap, Dallas Card, Saadia Gabriel, Yejin Choi, and Noah Smith. 2019. The risk of racial bias in hate speech detection. In *Proceedings of the 57th Annual Meeting of the Association for Computational Linguistics*, page 1668–1678. Association for Computational Linguistics.

Abu Bakr Soliman, Kareem Eissa, and Samhaa R. El-Beltagy. 2017. Aravec: A set of arabic word embedding models for use in arabic nlp. *Procedia Computer Science*, 117:256 – 265. Arabic Computational Linguistics.

Marcos Zampieri, Preslav Nakov, Sara Rosenthal, Pepa Atanasova, Georgi Karadzhov, Hamdy Mubarak, Leon Derczynski, Zeses Pitenis, and Çağrı Çöltekin. 2020. SemEval-2020 Task 12: Multilingual Offensive Language Identification in Social Media (OffensEval 2020). In *Proceedings of SemEval*.